\pdfoutput=1

\documentclass[11pt]{article}

\usepackage[preprint]{acl}

\usepackage{times}
\usepackage{latexsym}
\usepackage{multirow}

\usepackage[T1]{fontenc}

\usepackage[utf8]{inputenc}

\usepackage{microtype}

\usepackage{inconsolata}

\usepackage{graphicx}

\usepackage{amsmath}

\usepackage{epigraph}

\usepackage{booktabs}
\usepackage{multirow}
\usepackage{array}
\usepackage{colortbl}
\usepackage{xcolor}
\usepackage{caption}
\usepackage{subcaption}
\usepackage{adjustbox}
\usepackage{makecell}
\usepackage{siunitx}
\usepackage{cellspace}
\usepackage{float}
\usepackage{lscape}  %
\definecolor{green2}{rgb}{0.56, 0.93, 0.56}
\definecolor{blue2}{rgb}{0.56, 0.56, 0.93}

\usepackage{todonotes}
\usepackage{tabularx, booktabs}
\usepackage{pifont}
\usepackage{tcolorbox}

\title{KRISTEVA:  

Close Reading as a Novel Task for Benchmarking Interpretive Reasoning}

\author{
Peiqi Sui$^{1*}$\quad Juan Diego Rodriguez$^2$\quad Philippe Laban$^3$\\
\textbf{Dean Murphy}$^2$\quad \textbf{Joseph P. Dexter}$^4$\quad \textbf{Richard Jean So}$^1$\quad \textbf{Samuel Baker}$^2$\quad \textbf{Pramit Chaudhuri}$^2$\\
$^1$McGill University\quad 
$^2$UT Austin\quad 
$^3$Microsoft Research\quad 
$^4$University of Macau\\
\texttt{$^*$peiqi.sui@mail.mcgill.ca}
}

\newcommand{\finalnumberofquestions}{1,331 } %

\begin{document}
\maketitle
\begin{abstract}
Each year, tens of millions of essays are written and graded in college-level English courses. Students are asked to analyze literary and cultural texts through a process known as close reading, in which they gather textual details to formulate evidence-based arguments. Despite being viewed as a basis for critical thinking and widely adopted as a required element of university coursework, close reading has never been evaluated on large language models (LLMs), and multi-discipline benchmarks like MMLU do not include literature as a subject. To fill this gap, we present KRISTEVA, the first close reading benchmark\footnote{Our benchmark is publicly available on huggingface (\url{https://huggingface.co/datasets/McGill-NLP/KRISTEVA}).} for evaluating interpretive reasoning, consisting of 1331 multiple-choice questions adapted from classroom data. With KRISTEVA, we propose three progressively more difficult sets of tasks to approximate different elements of the close reading process, which we use to test how well LLMs may seem to understand and reason about literary works: 1) extracting stylistic features, 2) retrieving relevant contextual information from parametric knowledge, and 3) multi-hop reasoning between style and external contexts. Our baseline results find that, while state-of-the-art LLMs possess some college-level close reading competency (accuracy 49.7\% - 69.7\%), their performances still trail those of experienced human evaluators on 10 out of our 11 tasks.

\end{abstract}

\epigraph{``It is not surprising that the detailed analysis of metaphors… sometimes feels like extracting cube-roots in the head.''}{-- I.A. Richards, (\citeyear{richards1936philosophy})}

\section{Background}
Close reading is “the detailed analysis of the complex interrelations and ambiguities (multiple meanings) of the verbal and figurative components within a [literary] work” \cite[217]{abrams2009glossary}. As a uniquely text-centric form of interpretive reasoning, close reading methodologies posit that aesthetic choices about literary and cultural texts are not trivially subjective or arbitrary preferences. Instead, such methodologies pay meticulous attention to how the workings of language, form, and style generate meaning, rigorously observing, analyzing, and leveraging formal and stylistic features they present as textual evidence for falsifiable claims about literary or cultural texts.

\begin{figure}[t!]
    \centering
    \includegraphics[width=1\linewidth,trim=0mm 0mm 310mm 0mm,clip]{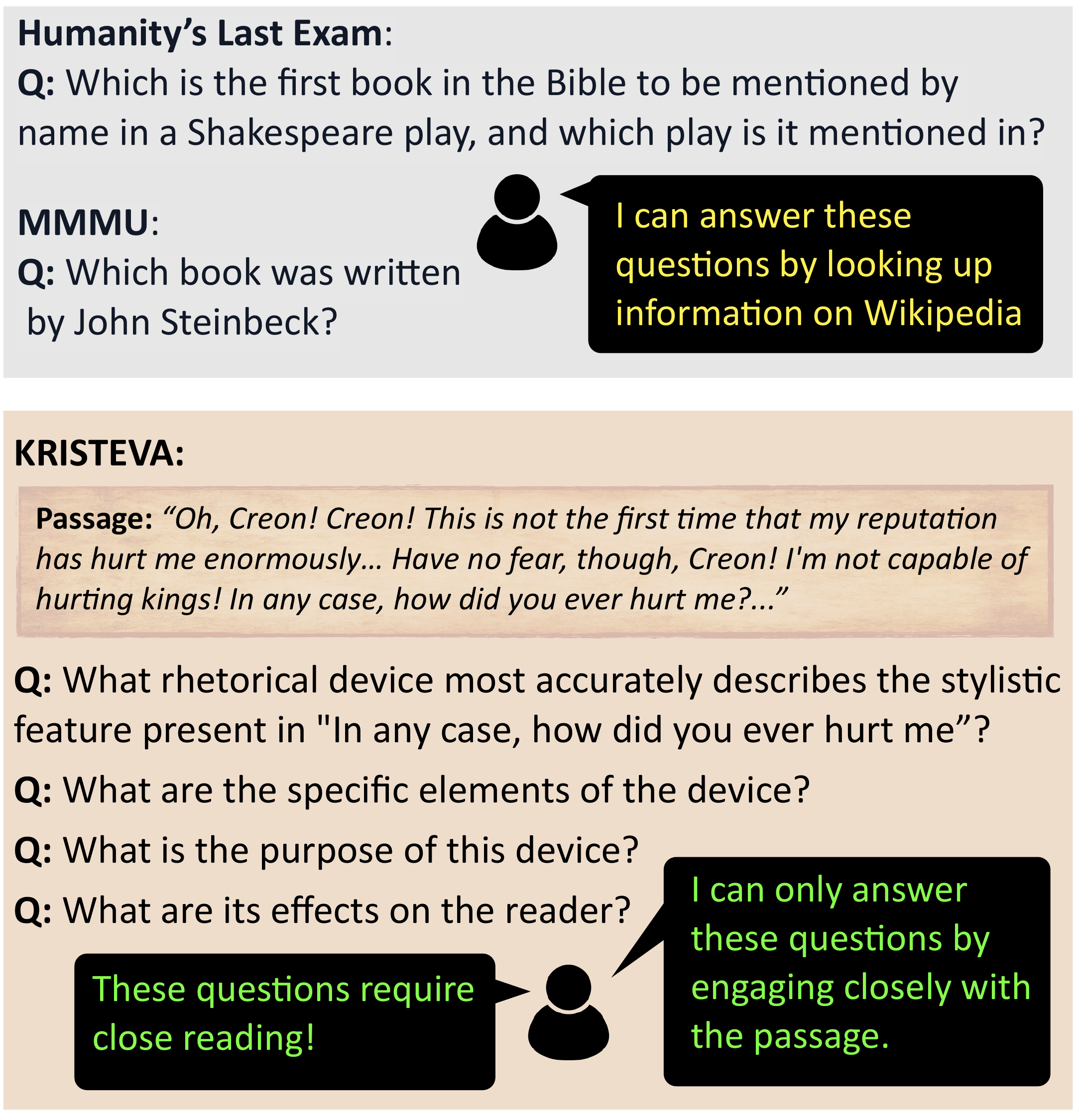}
    \caption{Examples of KRISTEVA questions that require interpretive reasoning to answer, compared to the purely informational literature questions from existing benchmarks.}
    \label{fig:kristeva_examples}
\end{figure}

As a skill, close reading has long been considered essential for cultivating critical thinking competencies that underpin active and informed participation in the civil discourse of participatory democracies \cite{dewey:how}. Recent studies argue that interpretive skills developed through close reading can be transferred to raising awareness of social justice \cite{hooks1994teaching}, recognizing disinformation and misinformation \cite{carillo2018teaching,mcgrew2020learning}, developing digital and media literacy \cite{hayles2010we,hobbs2010digital}, and fostering empathy \cite{charon2006narrative}. Reflecting its importance, close reading has been adopted both as a common requirement in college-level coursework \cite{bialostosky2006should} and as a pedagogical benchmark for secondary education \cite{CommonCore2012revised}. This policy trajectory aligns with predominant viewpoints within the humanities. Literary and cultural studies scholars argue that close reading enables individuals to better recognize and articulate patterns of political, social, and economic meaning \cite{levine2015forms}, thereby potentially converging on common understandings and reducing polarization. 
Evaluating LLMs for close reading not only measures their ability to interpret literature, but could also clarify what it means for such models to have the foundational skills in critical reasoning about complex social issues.  

However, no existing benchmark directly assesses LLMs on their ability to perform close reading. This omission is reflective of a broader gap in reasoning benchmarks—specifically, a lack of benchmark tasks where there may be no definitive “right” answer but certain responses can be deemed clearly wrong or unreasonable. Notably, very few benchmarks evaluate LLMs for aesthetic judgment \cite{hullman2023artificial}, a form of reasoning that requires negotiating a balance between components that are subjective (with no strictly correct answers) and objective (with demonstrably wrong answers). Some pioneering work in this area has explored the ability of LLMs to display an understanding of humor \cite{hessel-etal-2023-androids} or music \cite{yuan-etal-2024-chatmusician}. We applaud these efforts, and we aim to further expand the scope of LLM reasoning evaluation to include what we see as close reading's higher-order, synthetic reasoning and understanding: a task domain in which interpretations are not categorically correct or incorrect but can be judged on their plausibility for supporting arguments that a careful reader would find persuasive \cite{sinykin2025close}. Close reading thus exemplifies an omnipresent yet under‑studied class of reasoning challenges that hold relativist instead of positivistic ground truth.

To this end, we present KRISTEVA (\underline{\text{C}}lose \underline{\text{R}}eading and \underline{\text{I}}nterpretive Rea\underline{\text{s}}oning with \underline{\text{T}}extual \underline{\text{Ev}}idence), the first benchmark that evaluates LLMs for 1) close reading as a form of reasoning previously overlooked by the NLP community, 2) college-level knowledge in the literary domain, and 3) figurative language understanding as multi-hop reading comprehension (Figure \ref{fig:kristeva_examples}). KRISTEVA consists of \finalnumberofquestions multiple-choice questions extracted from college-level exam data, along with a novel task structure adapted from UT Austin’s \textit{Critical Reader’s Interpretive Toolkit} (CRIT),\footnote{\url{https://liberalarts.utexas.edu/english/the-critical-reader-s-toolkit.html}} a heuristic framework widely used to teach close reading at the college level. Our tasks are designed to evaluate how effectively LLMs can perform a sequence of intermediary analytical steps commonly presented in college literature classrooms as an essential pedagogical scaffold that guides students towards producing evidence-based literary interpretations. We find that, while competitive, numerous state-of-the-art LLMs still fall behind the top-line human performance of experienced close readers on these tasks.

\begin{figure*}[t]
    \centering
    \includegraphics[width=1\linewidth,trim=0mm 95mm 0mm 0mm,clip]{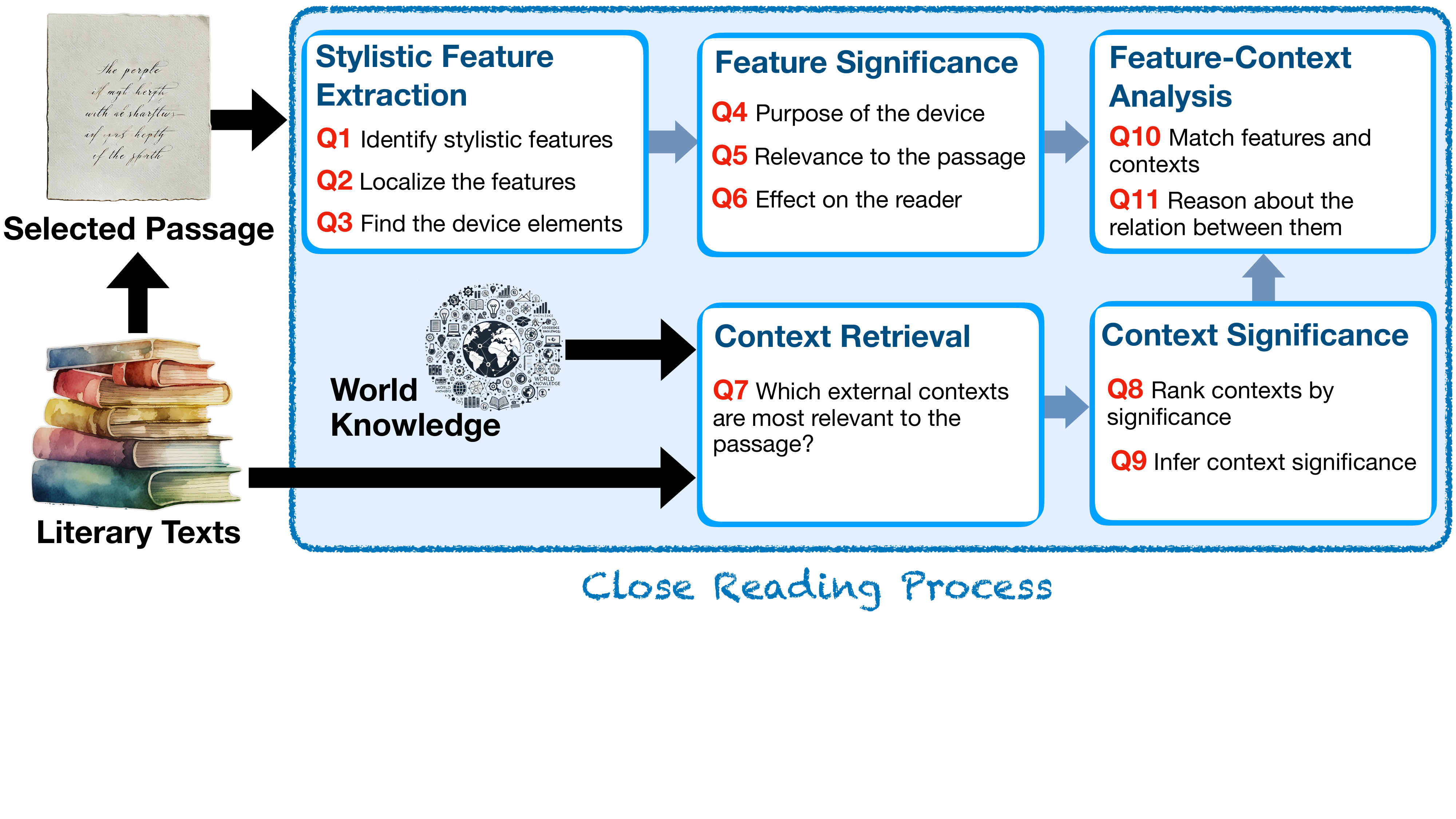}
    \caption{Question types in KRISTEVA correspond to distinct stages in the close reading process, involving both extractive tasks (e.g., stylistic feature extraction) and reasoning tasks (e.g., reasoning about the relation between features and relevant external context).} 
    \label{fig:tasks_flowchart}
\end{figure*}

For NLP, a particular strength of close reading as a data source is that it organically combines into a unified set of tasks two longstanding but isolated challenges of natural language understanding (NLU): figurative language understanding (FLU) \cite{chakrabarty-etal-2022-flute} and multi-hop reading comprehension \cite{welbl-etal-2018-constructing}. Our study is also an initial exploration of how the vast quantities of high-quality text data produced in the routine educational activities of humanities departments might be analyzed by and leveraged for NLP. 

\section{Task Description}
The KRISTEVA benchmark is adapted from CRIT, a heuristic framework developed by UT Austin’s English department for teaching close reading in literature courses required by UT’s undergraduate program. CRIT breaks close reading down into a step-by-step process: \textit{paraphrase}, \textit{observe}, \textit{contextualize}, \textit{analyze}, \textit{argue}, and \textit{reflect}. Each of its six sequential steps is guided by a set of exploratory questions that significantly reduces the cognitive load required for producing robust, evidence-based textual arguments. Pedagogically, humanities professors who use the CRIT tool have seen a clear positive effect on students' ability to perform more focused, detailed, and sustained textual analysis \cite{bares2020close}. 

To create KRISTEVA, we operationalize three of CRIT’s six steps into 11 distinct and sequential tasks designed for the evaluation of LLMs, each targeting a discrete cognitive procedure (Table \ref{tab:kristieva-tasks}). Seven out of the 11 tasks target novel forms of interpretive reasoning, in ascending order of complexity, that are essential building blocks for making a successful evidence-based interpretive argument (Figure \ref{fig:tasks_flowchart}). Detailed descriptions of each task, and of the CRIT steps to which they correspond, are presented in Appendix \ref{sec:tasks}. On a general level, they fall into three progressively more challenging clusters, as described in the following subsections. 

\paragraph{Q1-Q6: Stylistic Features}\quad Drawing on foundational theories of close reading, we define stylistic features as qualitative measures of a literary text's deviation from general domain language use. Writers often use expressions that have multiple, non-literal meanings and contradict standard, logical relationships; these departures can be justified on aesthetic grounds when they achieve particular effects rarely associated with functional language \cite{richards1929practical}. Consider, for instance, the difference between reading a technical manual and a poem: both require expertise on the part of the reader, but only one creates the expectation that multiple reasonable interpretations—potentially conflicting, or referring to entirely different phenomena—are possible. 
Stylistic features are knowledge-based representations of such patterns, 
like figurative language, sonic patterns, poetic form, diction, syntax, and narrative devices, that distinguish literary from non-literary texts.

KRISTEVA focuses on these features because of their importance to close reading, which has been described as a heightened sensitivity to nuanced textual elements and patterns \cite{guillory2025close}. Three tasks explicitly target the extraction and mapping of such features: detection (Q1), localization (Q2), and elaboration (Q3), along with a fourth task that reasons about the possible purpose of including these features in the passage (Q4). Close reading also entails the judgment of a work's literary merit—that is, determining whether it successfully leverages its stylistic features to conjure up a compelling enough effect that justifies the cognitive resources required to process these deviations from conventional language use. To evaluate this aspect of interpretive reasoning, two additional questions address a feature’s relative significance within its passage compared to other previously identified features (Q5),
and the specific effect it achieves for the reader (Q6). Overall, these tasks follow \citet{sravanthi-etal-2024-pub}'s framing of FLU as pragmatics capabilities, expanding the scope of the existing evaluations (Section \ref{sec:FLU}) to the interpretation of figurative language's affordances (Q4), significance (Q5), and effectiveness (Q6) as a form of communication.

\paragraph{Q7–Q9: Contextual Information}\quad We define context as the broader external circumstances within which a literary work is positioned, circumstances that might not be immanent within the text itself but are highly pertinent to its meaning. Plausible contexts include (but are not limited to) cultural, historical, literary,\footnote{Here ``literary'' factors include contextual information about literature as a field of social practice, such as what influences, generic conventions, or political constraints may have been relevant to the composition of the text.} and biographical factors that could help enrich the reader's understanding of the text. KRISTEVA evaluates models for both retrieving relevant contextual frames from parametric knowledge (Q7), and inferring their relative significance to the passage (Q8-Q9).

\paragraph{Q10–Q11: Multi‑hop Reasoning Between Features and Contexts}\quad Multi-hop reasoning requires chaining together multiple pieces of information, often across multiple documents or from external knowledge sources, to perform inference that cannot be derived from any single piece alone. In the case of close reading, multi-hop reasoning more specifically involves reasoning between a passage, features extracted from it, and contexts external to it. Since the deployment of stylistic features often involves managing trade-offs between language efficiency and the potential for emergent meaning, multi-hop reasoning is crucial for interpreting how the interplay between form and content can reveal novel semantic meaning not immediately available in the text itself. Such reasoning also drives aesthetic judgments in the literary domain, insofar as such judgments parse a text's context-grounded features in order to ascertain whether the outcome of such trade-offs renders a given text worthwhile. Although some earlier tasks could also be thought to involve multi-hop reasoning, Q10-11 directly require the combination of multiple pieces of contexts through matching features identified in previous questions to corresponding contexts (Q10), and inferring the most plausible connection between a given feature-context pair (Q11). %

\begin{figure}[t!]
    \centering
    \includegraphics[width=1\linewidth,trim=0mm 50mm 0mm 0mm,clip]{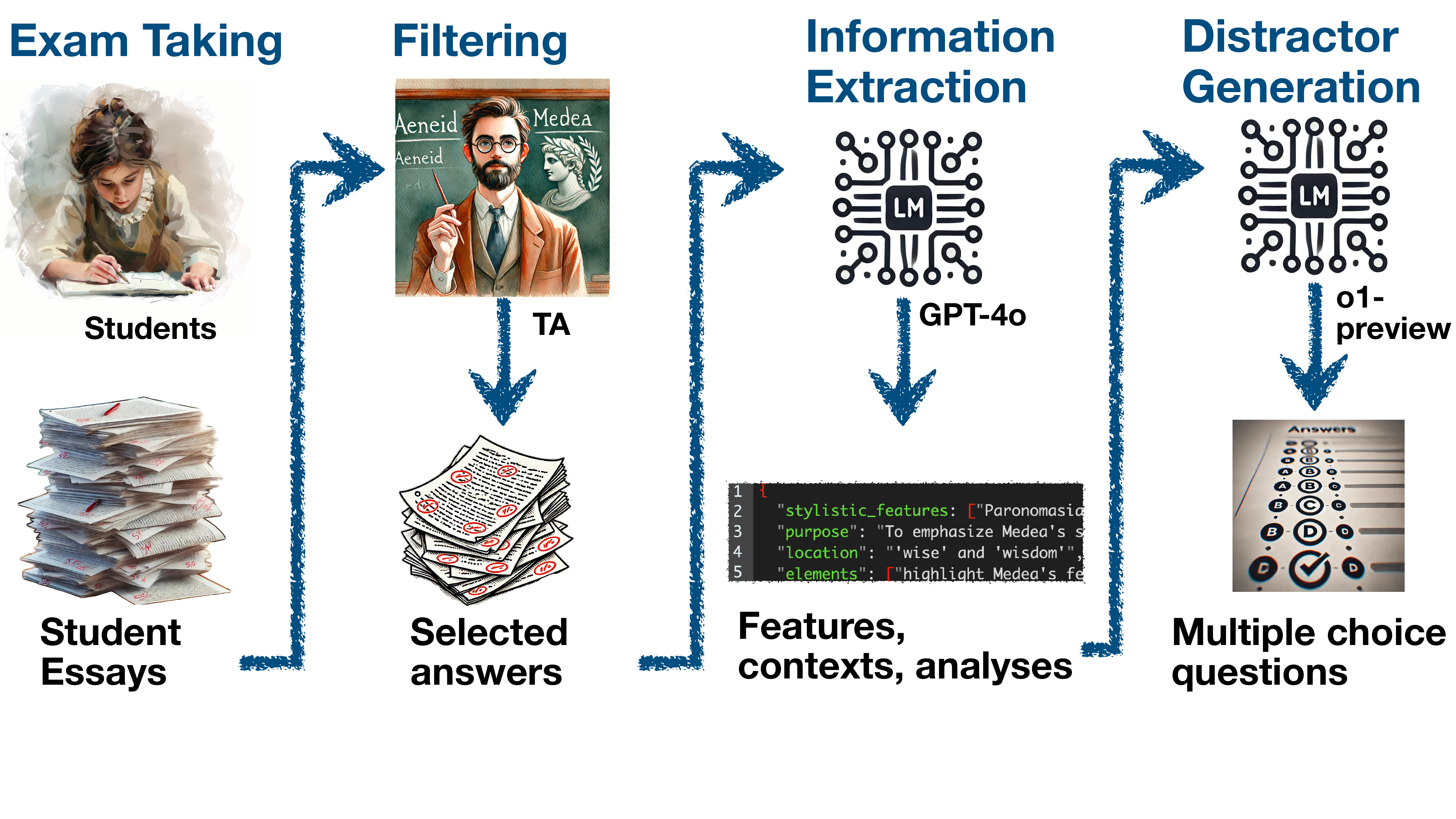}
    \caption{The dataset construction pipeline. Instructors manually filtered students' essays. We used GPT-4o to extract literary features from essays with the best answers. Finally, we used o1 to generate distractors (incorrect choices) for each multiple-choice question.}
    \label{fig:dataset_construction}
\end{figure}

\begin{table*}[t]
\small
\centering
\setlength{\tabcolsep}{3pt} %
\renewcommand{\arraystretch}{1.5} %

\begin{tabularx}{\textwidth}{p{4.5cm} p{4.5cm} p{3cm} c c} 
\toprule
\textbf{CRIT Step} & \textbf{KRISTEVA Tasks} & \textbf{In NLP Terms} & \textbf{Reasoning} & \textbf{Distractors} \\
\midrule

\multirow{6}{4.2cm}{\textbf{Observe}: Identify stylistic features from the passage and analyze their purpose and effect.} 
& Q1 Feature Type & \multirow{6}{=}{Figurative Language Understanding} &  & \ding{51} \\
& Q2 Feature Location &  &  & \ding{51} \\
& Q3 Feature Elements &  &  & \ding{51} \\
& Q4 Feature Purpose &  & \ding{51} & \ding{51} \\
& Q5 Feature Significance Ranking &  & \ding{51} &  \\
& Q6 Feature Significance Inference &  & \ding{51} & \ding{51} \\
\midrule

\multirow{3}{4.2cm}{\textbf{Contextualize}: Provide relevant pieces of cultural, historical, or literary contextual frames.} 
& Q7 Relevant Context Retrieval & Information Retrieval &  & \ding{51} \\
& Q8 Context Significance Ranking & \multirow{2}{=}{Multi-hop Reasoning} & \ding{51} &  \\
& Q9 Context Significance Inference &  & \ding{51} & \ding{51} \\
\midrule

\multirow{2}{4.2cm}{\textbf{Analyze}: Connect features with contexts and explain how they inform each other.} 
& Q10 Feature-Context Matching & \multirow{2}{=}{Multi-hop Reasoning} & \ding{51} &  \\
& Q11 Feature-Context Reasoning &  & \ding{51} & \ding{51} \\
\bottomrule
\end{tabularx}
\vspace{-1mm}
\caption{KRISTEVA task structure, adapted from UT Austin’s CRIT framework}
\label{tab:kristieva-tasks}
\end{table*}

\section{Dataset Construction}

\subsection{Data Collection}
\subsubsection{Data Source}
In educational settings, a minimal but sufficient implementation of CRIT's heuristic steps typically takes the form of a short analytical essay. As a pedagogical tool, each CRIT essay is designed as a self-contained unit of close reading, some assigned for practice and some for graded evaluation.

We collect 49 de-identified essays and grades from three exams in a university-level literature course that adapts the CRIT framework to help structure its close reading pedagogy. The course (``Introduction to Classical Literature: Forms, Cultures, Histories'') gives mostly first-year college students an introduction to world literature covering multiple historical periods, cultures, and genres. As an introductory course, it provides students with a foundational template for how to interpret literature. This template is partly based on the CRIT framework, although many of its elements are standard in the teaching of literary criticism. A major component of the course examinations is a single analytical essay focused on close-reading of a short literary passage drawn from a work students have previously studied.
\subsubsection{Data Processing}
The ground truth on data quality is directly established from the final grade the course instructor assigned to each essay. We leverage this grade to filter the collected essays and exclude low-score entries under 80\%. In addition, the course gives students the opportunity to revise and resubmit their first two exam essays for regrading, which in most cases result in significant improvements in data quality; when available, we replace the original exam essays with their revised versions. 

As discussed in Section 1, close reading differs from general domain reasoning tasks in that there are no objectively correct interpretations of literary works, but rather comparatively more or less reasonable ones. Critics evaluate interpretations on multiple bases often located in very different domains (e.g., how well grounded the interpretation is in a work's cultural context, or how it is received among some audience or other). This setting poses an epistemic challenge for validating the ground truth, since multiple answers could be correct at the same time.
To address this potential issue, we ask the instructor to perform a second manual check to ensure that for each correct answer it would also be possible to generate three distinct distractors that were less reasonable answers to the question.

\subsection{Benchmark Construction}
We build a pipeline that converts the unstructured texts from close reading essays into \finalnumberofquestions multiple-choice questions ready for LLM evaluation. The specific prompts we use for the multiple choice question (MCQ) construction pipeline can be found in Appendix \ref{appendix:benchmark-construction-prompts}.
\subsubsection{Question and Answer Extraction}
Structured representations of stylistic features, external contexts, and the connection between the two are first extracted from each essay, or summarized if the essay is too long. Some combination of this information, depending on the expected input and output of each task, is then used to construct each type of question and its corresponding answer. A detailed input/output schema of each question type is presented in Table \ref{tab:tasks-details}.

\subsubsection{Generation of Distractors}
While some MCQs simply reuse answers from earlier tasks as less significant options, others require the creation of entirely new distractors. We use o1-preview to generate three distractors for each of the 1,178 questions (7/11 question types) that require distractors (prompts shown in Appendix \ref{appendix:benchmark-construction-prompts}). Each distractor should closely mirror the structure and syntax of the correct answer to seem plausible and avoid confounders, while diverging semantically to present relatively less compelling interpretations. Distractor generation for MCQs is a challenging  problem \cite{stasaski2017multiple}. In order to ensure the quality of distractors, we experiment with the use of other LLMs to generate distractors (such as GPT-4o and Qwen) and perform manual inspection. Our inspection leads us to conclude that o1-preview generates on average the most relevant and challenging distractors for the close-reading literary domain.

Once distractors are generated for a question, the correct answer and distractors are merged into a final list of answer options, which is then shuffled, ensuring that there are no answer positional biases in the dataset. This shuffled list of answer options is presented in this arbitrary order in all the experiments we conduct.

\begin{table*}[t]
\small
\centering
\renewcommand{\tabcolsep}{1.5mm}
\begin{tabular}{lccccc|cccccccc|c}
\toprule
& \multicolumn{5}{c|}{\textbf{Non-reasoning}} & \multicolumn{8}{c}{\textbf{Reasoning}} & \textbf{Overall} \\
\cmidrule(r){2-6} \cmidrule(l){7-14} \cmidrule(l){15-15}
& \textbf{Q1} & \textbf{Q2} & \textbf{Q3} & \textbf{Q7} & \textbf{avg} 
& \textbf{Q4} & \textbf{Q5} & \textbf{Q6} & \textbf{Q8} & \textbf{Q9} & \textbf{Q10} & \textbf{Q11} & \textbf{avg} 
& \\
& \scriptsize{(209)} & \scriptsize{(144)} & \scriptsize{(209)} & \scriptsize{(139)} & 
& \scriptsize{(166)} & \scriptsize{(53)} & \scriptsize{(160)} & \scriptsize{(42)} & \scriptsize{(76)} & \scriptsize{(66)} & \scriptsize{(67)} & 
& \\
\midrule
Random         & 25.2 & 24.7 & 25.6 & 25.0 & 25.2 & 22.2 & 37.7 & 24.1 & 33.3 & 28.2 & 24.5 & 23.1 & 28.5 & 25.5 \\
\midrule
Qwen2.5-7B     & 34.4 & 91.7 & 75.1 & 55.4 & 62.3 & 64.5 & 58.5 & 73.1 & 33.3 & 75.0 & 27.3 & 82.1 & 60.7 & 62.9 \\
Qwen2.5-14B    & 40.7 & 95.8 & 76.6 & 54.7 & 65.2 & 66.3 & 58.5 & 71.3 & 28.6 & 77.6 & 34.8 & 82.1 & 61.7 & 64.8 \\
Qwen2.5-32B    & 47.4 & \cellcolor{green2}98.6 & \cellcolor{green2}83.3 & 62.6 & 71.4 & 66.9 & 60.4 & 70.0 & \cellcolor{green2}35.7 & 73.7 & 37.9 & 88.1 & 63.3 & 68.5 \\
OLMo-2--7B    & 38.3 & 84.7 & 57.4 & 49.6 & 55.6 & 37.3 & 43.4 & 56.9 & 31.0 & 53.9 & 31.8 & 61.2 & 44.3 & 51.3 \\
OLMo-2-13B    & 40.2 & 93.1 & 70.3 & 55.4 & 62.8 & 57.8 & 52.8 & 68.8 & 33.3 & 64.5 & 28.8 & 76.1 & 55.4 & 60.8 \\
OLMoE-1B-7B   & 29.7 & 81.2 & 52.6 & 46.0 & 50.2 & 46.4 & 49.1 & 54.4 & 23.8 & 64.5 & 27.3 & 62.7 & 48.2 & 49.7 \\
Gemma-2-2B    & 34.9 & 93.8 & 60.3 & 46.8 & 56.6 & 44.6 & 41.5 & 56.2 & 19.0 & 72.4 & 34.8 & 68.7 & 48.1 & 53.9 \\
Gemma-2-9B    & 41.6 & 94.4 & 72.2 & 58.3 & 64.7 & 65.1 & 58.5 & 69.4 & 31.0 & 77.6 & 37.9 & 83.6 & 62.0 & 64.4 \\
Gemma-2-27B   & 43.5 & 95.8 & 76.6 & 62.6 & 67.8 & 65.1 & 58.5 & 71.9 & 26.2 & 76.3 & 40.9 & 82.1 & 61.9 & 63.6 \\
Llama-3-8B    & 42.1 & 94.4 & 71.8 & 55.4 & 64.1 & 60.2 & 60.4 & 63.7 & 31.0 & 68.4 & 40.9 & 82.1 & 59.9 & 62.5 \\
Llama-3.1-8B  & 43.1 & 95.1 & 72.2 & 56.8 & 64.9 & 56.0 & 58.5 & 68.1 & 31.0 & 72.4 & 34.8 & 76.1 & 58.0 & 62.5 \\
Llama-3.1-70B & 46.9 & 96.5 & 78.9 & 54.7 & 67.8 & 63.9 & 60.4 & 70.6 & 31.0 & 71.1 & 36.4 & 88.1 & 61.9 & 66.0 \\
Llama-3.3-70B & 45.9 & 97.2 & 79.4 & 57.6 & 68.4 & 67.5 & 60.4 & 73.1 & \cellcolor{green2}35.7 & 68.4 & 37.9 & 89.6 & 63.2 & 67.2 \\
Phi-4         & \cellcolor{green2}49.3 & 97.9 & \cellcolor{green2}83.3 & 64.7 & \cellcolor{green2}72.2 & \cellcolor{green2}67.5 & \cellcolor{green2}62.3 & \cellcolor{green2}75.6 & \cellcolor{green2}35.7 & 76.3 & 37.9 & 83.6 & \cellcolor{green2}64.3 & \cellcolor{green2}69.7 \\
Mistral-7B    & 34.4 & 92.4 & 70.8 & 54.0 & 60.8 & 54.2 & 52.8 & 65.6 & \cellcolor{green2}35.7 & 63.2 & 33.3 & 74.6 & 54.6 & 59.1 \\
\midrule
GPT-4o-mini   & 45.0 & 94.4 & 78.0 & 62.6 & 68.3 & 61.5 & 58.5 & 73.8 & 33.3 & 75.0 & 40.9 & \cellcolor{green2}91.0 & 62.5 & 66.9 \\
GPT-4o        & 41.2 & 96.5 & 77.5 & 64.8 & 67.9 & 66.3 & 56.6 & \cellcolor{green2}75.6 & 33.3 & \cellcolor{green2}78.9 & 43.9 & 85.1 & 63.4 & 67.5 \\
o1-mini       & 43.5 & 95.8 & 77.0 & 54.0 & 66.0 & 57.8 & 60.4 & 68.1 & \cellcolor{green2}35.7 & 77.6 & 36.4 & 79.1 & 60.4 & 64.1 \\
o1-preview    & 40.7 & 97.2 & 74.6 & \cellcolor{green2}67.6 & 67.8 & 63.9 & 49.1 & 72.5 & \cellcolor{green2}35.7 & 77.6 & \cellcolor{green2}47.0 & \cellcolor{green2}91.0 & 61.5 & 66.8 \\
\midrule
Evaluator 1   & 43.5 & \cellcolor{blue2}100.0 & 52.2 & \cellcolor{blue2}72.2 & 63.7 & 63.6 & \cellcolor{blue2}75.0 & 61.1 & 0.0  & 77.8  & \cellcolor{blue2}71.4 & \cellcolor{blue2}100.0 & \cellcolor{blue2}68.7 & 65.4 \\
Evaluator 2   & \cellcolor{blue2}66.7 & \cellcolor{blue2}100.0 & \cellcolor{blue2}91.7 & \cellcolor{blue2}75.0 & \cellcolor{blue2}82.5 & \cellcolor{blue2}71.4 & 0.0 & 71.4 & \cellcolor{blue2}66.7 & \cellcolor{blue2}100.0 & 40.0 & 60.0 & 50.5 & \cellcolor{blue2}74.7 \\
Evaluator 3   & \cellcolor{blue2}65.2 & 94.1 & 69.6 & 64.0 & 72.0 & 52.9 & 0.0 & 50.0 & 28.6 & 70.0 & 41.7 & 66.7 & 39.0 & 61.5 \\
Weighted Average & \cellcolor{blue2}57.1 & 97.5 & 67.3 & \cellcolor{blue2}69.3 & 70.8 & 60.7 & 28.8 & 58.5 & 25.1 & \cellcolor{blue2}78.9 & \cellcolor{blue2}52.8 & 78.2 & 50.0 & 65.6 \\
\bottomrule
\end{tabular}
\caption{Performance (Acc) of LLMs on the KRISTEVA benchmark in zero-shot setting alongside a human baseline. We use \colorbox{green2}{green} to highlight the best model performance for each question type, and \colorbox{blue2}{blue} to highlight where human evaluators equal or outperform the best model. For all models, we report the direct match performance.%
The number of each type of questions is included in parentheses in the header.}
\label{tab:main-results}
\end{table*}

\section{Experimental Settings}
\subsection{LLMs}

We evaluate a range of language models—from 2B to 70B parameter models across various model families (Qwen, OLMo, Gemma, Llama, Phi, and Mistral), as well as GPT-4o and o1. We only use the instruction-tuned versions of the above models in a zero-shot setting\footnote{We omit chain-of-thought style evaluations because these have shown to mainly benefit mathematics and logic-related tasks \citep{to-cot-or-not}.}.

Each model is prompted to generate an answer in JSON format. We then extracte the answer and performed an exact match with the ground truth to assess accuracy; notably, no outputs were unparseable. The prompt used is provided in Appendix \ref{appendix:prompts}. %

All experiments are conducted with the Language Model Evaluation Harness \citep{eval-harness}\footnote{\url{https://github.com/EleutherAI/lm-evaluation-harness}, version 0.4.7} to ensure that our baseline results are reproducible.

\subsection{Human Evaluation}
To approximate a human baseline, we construct a subset of three unit tests, one for each exam by selecting MCQs from three essays per exam. We believe the evaluation results on the subset, which accounts for {percentage} of the dataset, are an unbiased estimate of the human performance over the whole benchmark.

We employ three experienced close readers (PhD students in the humanities) to answer these questions. Each evaluator completed one or two unit tests, with partial overlap to enable computation of inter-annotator agreement metrics (Section \ref{discussion}). Although most questions can be answered solely on the basis of the passage, we also provide the evaluators with the same class materials available to students to ensure subject-matter familiarity.

\section{Results}
Table \ref{tab:main-results} presents the performance of LLMs on the KRISTEVA benchmark, alongside a competitive human baseline. Phi-4 achieves the highest overall accuracy of 69.7, as well as the highest scores in both the reasoning (64.3) and non-reasoning (72.2) categories. Meanwhile, the o1-preview model, the largest reasoning model included in our experiment, stands out on most questions that require multi-hop reasoning. For most model families, larger variants generally outperform smaller ones (e.g., Qwen2.5-32B vs. Qwen2.5-14B), with the exception of Gemma-2-27B.

The top-line human performance surpasses the best-performing LLMs on 10 out of 11 tasks, generally by a wide margin. In addition, the best-performing human overall (evaluator 2) outperforms the best model (Phi-4) on 8 tasks. Notably, there is greater variability in performance among evaluators (average pairwise standard deviation of 29.3) compared to LLMs (5.47). In addition, human performance varies more across question types, with a coefficient of variation of 0.434 versus 0.322 for LLMs.

\section{Discussion}
\label{discussion}
\paragraph{Do LLMs outperform experienced humans on close reading?}\quad
While some models, most notably Phi-4, can approximate human-level performance on overall accuracy, the more fine-grained task-by-task breakdown shows that humans maintain a clear advantage in most aspects of close reading. The best performing models trail behind their human counterparts on 10 out of 11 tasks, eight of them more than 8\% in accuracy. This gap is likely even more pronounced in other evaluation settings, as our human baseline reported in Table \ref{tab:main-results} likely represents a conservative estimate: although our evaluators are experienced close readers, they still need a period of adjustment to KRISTEVA's MCQ format, which differs significantly from the actual practice of close reading. Two of the three evaluators reported considerable initial difficulty in cognitively adapting to the format of the questions, while another believed that time constraints limited their performance. These factors suggest that under more natural, open-ended evaluation settings of close reading, human performance would likely be even higher.

\paragraph{Do human evaluators agree with each other?}\quad
To assess the consistency of human judgments, we computed Krippendorff’s \(\alpha\) for evaluator pairs overlapping on the same unit tests. Agreement scores range from 0.523 (unit test 3) to 0.644 (unit test 2), indicating moderate consensus. Notably, the lower agreement score corresponds to evaluators from different departments (English and Classics), while evaluators from the same department (Classics) display higher agreement.

In contrast to the relative consistency observed among LLMs, human performance exhibits more significant variability. For some question types, one evaluator achieved high accuracy while another scored zero (Q5, Q8). It is important to acknowledge that such differences may simply be the result of the small sample size (i.e., only three evaluators). We hypothesize, however, that these discrepancies could also be influenced by domain expertise. The English literature PhD student (evaluator 1) excels on the more complex reasoning questions, perhaps due to greater familiarity with the CRIT framework or the theory of criticism underlying the approach. Meanwhile, the two Classics PhD students, who may have greater subject-matter familiarity with the particular types of literary texts or their historical circumstances, perform better on the extraction-based questions, as well as on the specific reasoning tasks involving external contexts. 
Our results suggest that, in small-sample settings, diversity in academic backgrounds and areas of specialization may drive volatility in human performance—a factor that future work should consider when defining human baselines for close reading.

\paragraph{What makes an LLM good at close reading?}\quad
 Although the best performing Phi-4 is a smaller model (14B), its high-quality, textbook-based training data might have a closer affinity to the college classroom data source from which KRISTEVA is collected. While larger models generally outperform their smaller variants, most tasks exhibit a more significant gap between Phi-4 and much larger models like Llama-3.1-70B. This difference suggests that data quality could be a more significant factor for interpretive reasoning ability than model scale, which further supports our call to explore the scalability of ethical data collection from college classroom settings.

 Despite outperforming on the three out of four tasks that require multi-hop reasoning (Q8, Q10, Q11), reasoning models like o1-preview do not exhibit any advantage in most tasks. This result is consistent with the findings of recent studies that chain-of-thought mainly improves mathematics and logic-related tasks \cite{to-cot-or-not}, while having a very limited impact on commonsense, knowledge, and soft reasoning tasks that are more relevant to the setting of KRISTEVA.

\section{Related Work}

\paragraph{College and PhD-level LLM Evaluations}\quad
Since OpenAI’s popularization of the term, “PhD-level intelligence” has rapidly caught on in the public discourse of AI as a tangible signpost for artificial general intelligence (AGI). Building on earlier general LLM evaluations at the college (MMLU) and graduate-levels of reasoning \cite{rein2023gpqa,sawada2023arb}, subsequent efforts have introduced domain-specific assessments in mathematics \cite{liu-etal-2024-mathbench,tsoukalas2024putnambench}, computer science \cite{song2024cs}, biology \cite{laurent2024lab}, history \cite{hauser2024large}, and psychology \cite{zhang2024conceptpsy}. However, very few of these single or multidisciplinary benchmarks includes literature as a test subject\footnote{Examples include ``Humanity's Last Exam'' \cite{phan2025humanity}, which contains eight questions on English literature and 15 on poetry, and MMMU. MMMU \cite{yue2024mmmu} also technically has a ``literature'' category, but most questions listed in that category are purely informational, concerning book covers and illustrations rather than the literary text itself (Figure~\ref{fig:kristeva_examples}). Some other benchmarks also concern literature, but exclusively in the Chinese language \cite{li-etal-2024-cmmlu,cao2024wenmind}.}—a surprising omission given OpenAI’s own results, which indicate that ChatGPT, GPT-4, and o1 all significantly underperform on AP English tests compared to other AP exams.\footnote{\url{https://openai.com/index/learning-to-reason-with-llms/}} The causes of this discrepancy have not yet been explored. We introduce KRISTEVA to begin to address this gap.

\paragraph{Multi-hop Reading Comprehension (MRC)}\quad
Since CosmosQA \cite{huang-etal-2019-cosmos}, there has been a growing interest in the evaluation of deeper reading comprehension capabilities that require reasoning components to extend beyond the literal understanding of the text \cite{dua-etal-2019-drop,sun-etal-2019-dream,yu2020reclor}. Evaluating such capabilities departs from earlier MRC benchmarking efforts that do not require reasoning \cite{rajpurkar-etal-2016-squad,chen-etal-2016-thorough}, or where the involvement of reasoning might even lead to a drop in performance \cite{jia-liang-2017-adversarial}. A hallmark of multi-hop MRC is its reliance on external information, either explicitly provided across multiple documents \cite{welbl-etal-2018-constructing} or implicitly elicited via common sense \cite{huang-etal-2019-cosmos}, to fully understand the passage at hand. Incorporating reasoning-based MRC into domain-specific continued pre-training has been shown to enhance performance both within specialized domains and on general benchmarks \cite{cheng2023adapting}.

Recent benchmarks further align MRC with more complex reasoning tasks, such as natural language inference (NLI) \cite{liu2023logiqa}, deep text understanding \cite{yao-etal-2023-korc}, critical reasoning \cite{kawabata-sugawara-2023-evaluating}, and extractive question answering \cite{basmov2024llms}. In keeping with this research direction, we formulate context-dependent close reading as a uniquely challenging form of multi-hop MRC: to successfully reason between literary form and content (Q9), models must first correctly extract from the passage both components of the logical connection: stylistic features and external context. To the best of our knowledge, ours is the first MRC benchmark to be based on the challenging domain of literary texts and to require reasoning on figurative elements. Additionally, our dataset is sourced from college-level long-range documents (essays), which offer higher volumes and text quality compared to the standardized testing venues of existing benchmarks, like Chinese ESL \cite{sun-etal-2019-dream} and LSAT \cite{yu2020reclor}.

\paragraph{Figurative Language Understanding (FLU)}\quad
\label{sec:FLU}
Initial benchmarks have evaluated FLU through QA \cite{rakshit-flanigan-2022-figurativeqa} and NLI \cite{stowe-etal-2022-impli}. Moving beyond simpler tasks like metaphor detection, more recent studies have approached FLU as a form of reasoning. However, the scope of their reasoning tasks remains limited to rationales \cite{chakrabarty-etal-2022-flute}, i.e., why something is a metaphor; explanations \cite{liu-etal-2022-testing,comsa-etal-2022-miqa}, i.e., what the metaphor means and a breakdown of its implications; or literal rewordings \cite{tong-etal-2024-metaphor}. Few frame FLU as a pragmatics capability \cite{sravanthi-etal-2024-pub}, and none require models to interpret figurative language’s broader significance (as when our Q4 and Q6 investigate whether a given metaphor is needed, what its representational affordances might be, and how the passage would be different without it), or judge its relative effectiveness (as when our Q5 inquires whether one metaphor might be considered as more successful than another).

In addition, most existing benchmarks address FLU at the sentence-level, with far less focus on figurative language in context \cite{chakrabarty-etal-2022-rocket}. These cleanly parsed datasets do not align with the real-world complexity of figurative language as cognitive processes—metaphors, for instance, rarely exist in isolation, but are embedded in larger bodies of surrounding texts that often themselves function as part of broader networks of figural causations \cite{auerbach:mimesis,white1999figural}. Consequently, KRISTEVA introduces more complex reasoning tasks with purpose, effect, and context that are necessary for understanding how figurative language contributes to a passage’s overall meaning or the literary work's narrative flow. As far as we know, KRISTEVA is the first benchmark to explicitly formulate FLU as a multi-hop reasoning task situated in the framework of multi-hop MRC.

\paragraph{Literary NLP}\quad The benchmark gap for close reading, the gold standard of evidentiary claims in literary studies, limits the advancement of NLP research in the literary domain, where the performance of general-domain NLP models tend to “drop precipitously” \cite{bamman-etal-2019-annotated}. Despite the utilization of literary corpora for tasks such as event extraction \cite{sims-etal-2019-literary}, information retrieval \cite{thai-etal-2022-relic}, and pretraining data detection \cite{chang-etal-2023-speak}, the literary domain remains relatively overlooked within the broader NLP community. While part of this challenge is inherent in the semantic ambiguity and pragmatic ineffability of literature, the development of literary NLP is more directly constrained by the bottleneck of standardized benchmark tasks, expert-annotated datasets, and generalizable evaluation metrics.

\section{Conclusion and Future Work}
We present KRISTEVA, the first close reading benchmark that evaluates the interpretive reasoning abilities of LLMs, featuring a novel task structure and a competitive human baseline. On comprehensive experiments with 19 models, we show that close reasoning presents several challenging tasks, and that LLMs still lag behind human performance.

Beyond the literary domain, the interpretive reasoning evaluated by KRISTEVA could be applicable to a range of NLP tasks. The close reading skills that KRISTEVA quantifies can serve as a proxy for evaluating long-range document understanding and help enhance the hermeneutic capabilities of LLMs. Additionally, close reading hones the reasoning ability to recognize narrative fidelity and coherence, which often serve as pragmatics protecting against harmful confabulation in human-to-human interaction; likewise, close reading abilities could potentially provide cultural guardrails against LLM confabulation \cite{sui-etal-2024-confabulation}. The ability to reason across latent stylistic and contextual spaces, addressed by Q10-Q11 in particular, also underpins the potential political affordances of close reading, for instance in the cases of disinformation recognition and media literacy. Moreover, KRISTEVA's emphasis on evidence-based interpretive judgment could enrich style transfer and other human-centered creative NLP domains by facilitating more robust and justified preference judgments. We plan to continue updating KRISTEVA to better address these evolving research needs.

As the first benchmark of its type, KRISTEVA follows the NLP community’s common use of MCQ format. Our subsequent work seeks to broaden the benchmark to open-ended evaluations of LLM free-text responses when directly prompted to perform close reading.

\section*{Limitations}

\subsection{Data}
Our current data source and collection process has several limitations. First, the scope of the study is limited to the classroom data of one course. Second, the course is at the introductory level, which limited the quality of the close readings gathered; its exams also do not implement the full CRIT (foregoing the ``argument'' step, which would have extended the length of the exam beyond the time available). Third, although two out of the three primary texts on the exams are translated from other languages, this work is performed entirely on texts in English.

\subsection{Human Evaluation}
The human evaluators who produced our current human baseline may have been disadvantaged by KRISTEVA's MCQ format, as discussed in Section \ref{discussion}. This effect possibly reflects the fact that its MCQs contain LLM-generated distractors, which could bias the benchmark towards being more LLM-solvable. Both factors may be emblematic of the more general problem of complex and partly subjective reasoning tasks being adapted to typical approaches to AI evaluation \cite{crawford2021atlas}. %
To address these disadvantages, in our future work we plan to explore both human-annotated distractors and open-ended evaluations of close readings.
\subsubsection{``Expertise''}
The term ``domain expertise'' is used in this paper to refer to literary critical competency at the level of a graduate student at a major research institution. Recent high-profile research has used a similar standard; for instance, the Ithaca tool for reconstructing damaged inscriptional texts was benchmarked against human annotations by ``graduate students of ancient history, with 7 years of historical and linguistic training and specializing in Greek history and epigraphic documents'' \cite{Assael2022}. As acknowledged by the Ithaca researchers, even experts at this level are ``not yet equivalent to (the very small number) of established specialists in the field.'' A further challenge for close reading of literary texts is the frequent need to draw on expertise beyond a single humanistic domain. %
Our results already suggest, for instance, potential differences in evaluation connected to disciplinary background. In addition, annotators had insufficient time to read through the full course materials, hence they focused their preparation primarily on the works from which the relevant close-reading passages were drawn. Future research, therefore, might fruitfully explore the impact of greater opportunities for annotators to prepare, different forms of preparation, and a larger number of annotators from varied disciplinary backgrounds. 

\section*{Ethics Statement}
The study protocol was submitted to the Institutional Review Board (IRB) at the UT Austin (ID: STUDY00006946). 
The IRB determined that this protocol meets the criteria for exemption from IRB review under 45 CFR 46.104 (1) Educational settings. The protocol required that the course data was de-identified by the course instructors before being shared with other members of the research team. Students consenting to participate in the study received \$20 in cash. Graduate research assistants were paid \$30 per hour for their contributions to the human evaluation data.

\section*{Acknowledgments}
This work was supported by Good Systems, a research grand challenge at UT Austin. CRIT, which helped inspire the format for this research, was developed in the Department of English at UT Austin by Professors Phillip Barrish, Evan Carton, Coleman Hutchison, and Frank Whigham, and PhD students Sydney Bufkin, Jessica Goudeau, and Jennifer Sapio. CRIT is a product of a Course Transformation Grant generously funded by the Office of the Executive Vice President and Provost. CRIT is licensed under a Creative Commons Attribution-NonCommercial 4.0 International License.

We thank Ziang Xiao for his valuable feedback on the experimental design of our benchmark. Finally, we thank our anonymous reviewers for their generous time and attention.

\bibliography{custom}

\appendix

\section{Humanistic Mission Statement}
Automating close reading is not an end goal of this work: for many reasons, including the basic fact that close reading is a practice whose significance largely derives from its status as a method of personal deliberation. However, developing computational models for close reading could help clarify and facilitate what we humans do when we engage in this practice, describe it, and teach it. For example, machine learning could enable new kinds of systematic comparisons among critical and literary-theoretical approaches.

Such comparisons could inform practitioners of literary criticism about a range of issues relating to their craft: about what kinds of task are prone to error, where human creativity excels, how disciplinary conventions shape analysis, and which skills warrant particular attention in our pedagogy. At the same time, the field of literary studies, with its distinctively subjective and associative forms of reasoning, can provide resources for the development of the next generation of language models, and a crucial test for them. Here as elsewhere, the craft of traditional disciplines has much to offer computational research \cite{underwood2019distant}. 

In the course of connecting these fields, we try to avoid attributing ``understanding,'' ``judgment,'' or other forms of critical consciousness to LLMs. When such attributions do happen, we consider them to make the same kind of sense as commonly seen attributions of human qualities to traditional human-created works, like books, or latent author figures composed from our experience of such works: as when we say that Coleridge's \textit{Lectures on Shakespeare} change our understanding of the Bard, ``himself a nature humanized, a genial understanding directing self-consciously a power and an implicit wisdom deeper even than our consciousness'' \cite{coleridge1849notes}.

\section{KRISTEVA Tasks Details}
\label{sec:tasks}
Please see Table \ref{tab:tasks-details} for the full input/output schema of KRISTEVA's task structure.

\begin{table*}[t]
\small
\centering
\setlength{\tabcolsep}{3pt} %
\vspace{-1mm}
\renewcommand{\arraystretch}{1.5} %

\begin{tabularx}{\textwidth}{p{3cm} X X} 
\toprule
\textbf{Question} & \textbf{Input / Output} & \textbf{Full Question} \\
\midrule

Q1 Feature Type & Input: \{location\} \newline Output: \{feature\_type\} & What rhetorical device is present in \{location\}? \\
Q2 Feature Location & Input: \{feature\_type\} \newline Output: \{location\} &  \\
Q3 Feature Elements & Input: \{feature\_type\}, \{location\} \newline Output: \{feature\_elements\} & In the \{feature\_type\} that occurs in \{location\}, what are the specific elements of the device? \\
Q4 Feature Purpose & Input: \{feature\_type\}, \{location\}, \{feature\_elements\} \newline Output: \{purpose\} & In the \{feature\_type\} that occurs in \{location\}, \{feature\_elements\}. What is the purpose of this device? \\
Q5 Feature Relative Importance & Input: \{features\} (answers of Q1-4) \newline Output: \{significant\_feature\} & Which of the following stylistic features is the most significant to the passage? \\
Q6 Feature Significance & Input: \{significant\_feature\} \newline Output: \{feature\_significance\} & In the \{feature\_type\} that occurs in \{location\}, \{feature\_elements\}. Which of the following best describes the significance of this device, and what are its effects on the reader? \\
Q7 Relevant Context Retrieval & Input: \{passage\} \newline Output: \{context\_type\}, \{context\_elements\} & Which external context is the most relevant to the following passage? \\
Q8 Context Relative Importance & Input: \{contexts\} (answers of Q7) \newline Output: \{significant\_context\}, \{context\_elements\} & Which of the following contextual information is the most significant to the passage? \\
Q9 Context Significance & Input: \{significant\_context\} \newline Output: \{context\_significance\} & In the \{context\_type\} that occurs in \{location\}, \{context\_elements\}. Which of the following best describes the significance of this device, and what are its effects on the reader? \\
Q10 Feature-Context Matching & Input: \{context\_type\}, \{context\_elements\}, \{features\} \newline Output: \{corresponding\_feature\} & Please identify the stylistic feature that the following \{context\_type\} information best helps to contextualize: \{context\_elements\}. \\
Q11 Feature-Context Reasoning & Input: \{context\_type\}, \{corresponding\_feature\}, \{selected\_passage\} \newline Output: Rationale for the \{feature\_context\_pair\} & What is the most reasonable connection between the \{context\_type\} context and the use of \{feature\_type\} feature in the following passage? \\

\bottomrule
\end{tabularx}
\caption{KRISTEVA task structure and question formats, adapted from UT Austin’s CRIT framework.}
\label{tab:tasks-details}
\end{table*}

\subsection{CRIT Step: Observe}
In this step of the CRIT framework that helped formalize the close reading pedagogy this study engages, students are asked to perform literary pattern recognition (So and Long, 2016): the observation of how the passage stands out as a literary one through its deviation from normal everyday language use, identified as features of form and style that help a passage accomplish its literary aims. These stylistic features, including figurative language, sonic patterns, poetic form, diction and syntax, narrative devices, etc., are widely considered to be characteristic of literary texts and distinguish them as a unique domain. 
Operationally, the “observe” step could be broken down into two components: 1) listing and explicating potentially significant stylistic features in the selected passage, and 2) inferring the purpose they serve in the passage and evaluate their rhetorical success in terms of what unique effect they have achieved (i.e., hypothesizing how the text would read differently if a given feature were removed).

\paragraph{Q1 (Feature Type), Q2 (Feature Location), Q3 (Feature Elements)}\quad 
Mostly following the existing framework of FLU, these three tasks collectively evaluate LLMs’ ability to accurately detect, locate, and explain figurative language and other stylistic features embedded in a given passage of literary text. Q1 asks models to detect the type of features present in a given passage (e.g., metaphor, alliteration, or symbolism). Q2 is the reversal of Q1, where models are given a feature type and asked to identify the part of the passage where it occurs. Q3 builds on the answers of Q1 and Q2 to require a higher level of stylistic feature understanding, prompting models to describe the specific elements of a given feature and location. 

For student essays that do not provide a specific location through direct quotes, Q1 and Q2 are collapsed into one question: ``Which of the following is the most prominent use of rhetorical device in the passage?''

Q1 and Q3 are standard figurative language identification tasks that structurally align with extant forms of evaluation in FLU. Q2, however, represents a departure by testing for open stylistic feature extraction from unstructured text: existing work in FLU tends to represent figurative language with “subject-relation-object” triples \cite{liu-etal-2022-testing} in clean, information-extraction style datasets. In contrast, complex rhetorical devices in literary texts (e.g, a Homeric simile) are often not neatly separable from the surrounding language. This scenario entails a more challenging form of FLU beyond only identifying and explaining the component of a device, requiring models to also parse out the language around it that aids in its construction.

\paragraph{Q4 (Feature Purpose), Q5 (Feature Significance Ranking), and Q6 (Feature Significance Inference)}\quad
Given a fully extracted and localized stylistic feature (outputs of Q2–Q4), Q5 asks the model to interpret why a particular device is used and how it influences the text. This question concerns the functional role of a given feature within a text (e.g., is it creating suspense, emphasizing an emotional tone, making a certain imagery more concrete, etc.) and its rhetorical success (i.e., what effect it could have on the reader’s interpretation or emotional response). Q5 \& Q6 builds on this analysis of effect to evaluate the model’s ability to weigh the relative importance of features. Specifically, it asks the model to compare the influence of multiple stylistic features (all outputs of Q1–Q4) within the selected passage, and rank them in descending order of significance by assessing their effects on the reader (how well they direct reader engagement and manage their attention and expectations). 
Together, Q4, Q5, and Q6 push the reasoning components of FLU to interpretations regarding figurative language’s broader significance – \textit{what they are for}, and \textit{why certain features are especially impactful} for the text.

\subsection{CRIT Step: Contextualize}
This step expands the interpretive scope of close reading by introducing elements external to the literary passage: pertinent configurations of facts and boarder circumstances, be it literary, biographical, cultural, or historical, that could serve as contextual frames for the passage and contribute to its overall meaning. Students are asked to draw on either their own world knowledge or course materials to extract a list of relevant contextual information, such as cultural artefacts, intellectual history, authorial influences, or wider societal developments, and consider how these external factors could be potentially significant for the passage.

\paragraph{Q7 (Relevant Context Retrieval)}\quad 
To evaluate the ``contextualize'' step, Q7 requires models to identify the most relevant piece of contextual information to a literary passage, and classify its type (literary, biographical, cultural, or historical). It tests if the model could effectively leverage its parametric knowledge to perform zero-shot world knowledge retrieval, and understand the relative significance of what it retrieved.

\section{Prompts}
\label{appendix:prompts}

\subsection{Benchmark Construction Prompts}
\label{appendix:benchmark-construction-prompts}

The following prompts were used to extract structured representations from CRIT essays and generate distractors:

\onecolumn
\begin{tcolorbox}[colback=gray!10, colframe=black, boxrule=0.5pt]
\scriptsize
\textbf{You are given the response to an exam that had four questions (Question 1-4) that analyze the following passage:}
\begin{quote}
[passage]
\end{quote}
Your goal is to extract the individual answers to each of the questions.

\textbf{Rules:}
\begin{itemize}
    \item If there are errors due to text extraction, such as excessive new lines, you can fix that.
    \item You can fix clear and unambiguous typoes.
    \item Your response should be a JSON object, with the following schema:
\end{itemize}

\texttt{"q1\_observe": [\{"id": 1, "type": "...", "location": "...", "elements": "...", "purpose": "..." \}, ...]} \\
\textit{(list of dictionaries, each with the following keys: id, type, location, elements, purpose)}

\begin{itemize}
    \item The \texttt{id} corresponds to a unique number for each rhetorical device (1, 2, 3, etc.) so it can be identified in the response.
    \item The \texttt{type} corresponds to a type of rhetorical device, such as: Allusion, Metaphor, Personification, etc.
    \item The \texttt{location} corresponds to an exact excerpt from the selected passage that corresponds to the rhetorical device. Include only what the student presents in direct quotations from the literary passage. If the student does not present any direct quotations, leave this key blank.
    \item The \texttt{elements} corresponds to the rhetorical device's components. In one sentence, simply outline the elements of what is being described as what (if the device is a metaphor, the elements are the tenor and the vehicle, etc.), or the specific language that carries the device (if the device is an alliteration, the elements are the letters that are alliterated, etc.). Some students might also offer an explanation for how the device's components differ from their literal meaning. If so, include the explanation as a second sentence.
    \item The \texttt{purpose} is the purpose and effect of the rhetorical device, extracted from the response. For instance, the purpose of \textit{"Word choice: words such as 'gnawing', 'rage' and 'crying' create an eerie and dark tone that fits throughout the passage"} is \textit{"to create an eerie and dark tone that fits throughout the passage"}. Please summarize the purpose and effect in one sentence if the student's description is too long. Some responses could be excessively short and not contain a purpose, e.g., \texttt{"epithets ('gem of Raghus', 'best of Raghus')"}, in which case, leave the purpose blank.
    \item You must only extract rhetorical devices listed in the response and not invent ones that are not present. You should only extract elements from the response for the elements/purpose, and not infer or create new ones.
\end{itemize}

\texttt{"q2\_context": [\{"id": 1, "type": "...", "elements": "..." \}, ...]} \\
\textit{(list of dictionaries, each with the following keys: id, type, elements)}

\begin{itemize}
    \item The \texttt{id} corresponds to a unique number for each context element (1, 2, 3, etc.) so it can be identified in the response.
    \item The \texttt{type} corresponds to a type of contextual element, such as: Historical, Cultural, Biographical, etc.
    \item The \texttt{elements} are a high-level description of the contextual element that is relevant to the selected passage, extracted from the response of the student. Focus on the factual information about the context external to the literary work. Please summarize the content in one sentence if the student's response is too long.
\end{itemize}

\texttt{"q3\_analyze\_i": [\{"id": 1, "type": "...", "corresponding\_id": 1, "significance": "..." \}, ...]} \\
\textit{(list of dictionaries, each with the following keys: id, type, corresponding\_id, significance)}

\begin{itemize}
    \item The \texttt{id} corresponds to a unique number for each significance (1, 2, 3, etc.) so it can be identified in the response.
    \item The \texttt{type} describes if the significance is about a rhetorical device or a contextual element. Return \texttt{"feature"} for rhetorical devices and \texttt{"context"} for contextual elements.
    \item The \texttt{corresponding\_id} is the id of the rhetorical device or contextual element that the significance corresponds to, drawn from the \texttt{id} of either \texttt{"q1\_observe"} or \texttt{"q2\_context"}.
    \item The \texttt{significance} is the significance of a rhetorical device or contextual element, extracted from the response. Please summarize the significance in one sentence if the student's description is too long.
\end{itemize}

\texttt{"q4\_analyze\_ii": [\{"feature\_context\_pair": "...", "feature\_id": 1, "context\_id": 1, "feature\_context\_conn": "..." \}, ...]} \\
\textit{(list of dictionaries, each with the following keys: feature\_context\_pair, feature\_id, context\_id, feature\_context\_conn)}

\begin{itemize}
    \item The \texttt{feature\_context\_pair} corresponds to the rhetorical device and the context connected together. Each pair should be listed as a string with the following structure: \texttt{"\{feature\_type\}, \{feature\_location\}; \{context\_type\}, \{context\_elements\}"}.
    \item The \texttt{feature\_id} corresponds to the id of the pair's rhetorical device identified in \texttt{"q1\_observe"}. If multiple rhetorical devices are described in the response, only select the first one. If no rhetorical device is described, make it -1.
    \item The \texttt{context\_id} corresponds to the id of the pair's context identified in \texttt{"q2\_context"}. If multiple rhetorical devices are described in the response, only select the first one. If no rhetorical device is described, make it -1.
    \item The \texttt{feature\_context\_conn} is a description in the response of how the rhetorical device is connected to its corresponding contextual element and what makes this connection significant.
\end{itemize}

\textbf{Overall, the JSON response you produce should therefore adhere to the following schema:}

\texttt{\{"q1\_observe": [...], "q2\_context": [...], "q3\_analyze\_i": [...], "q4\_analyze\_ii": [...]\}}

\textbf{Now proceed with the extraction for the following response:}

\end{tcolorbox}

\begin{tcolorbox}[colback=gray!10, colframe=black, boxrule=0.5pt]
\scriptsize
\textbf{You are given a JSON array of question objects, each containing the following fields:}
\begin{itemize}
    \item \texttt{"question\_number"} (string; one of: \texttt{"Q1"}, \texttt{"Q2"}, \texttt{"Q4"}, \texttt{"Q5"}, \texttt{"Q7"}, \texttt{"Q9"}, or \texttt{"Q11"})
    \item \texttt{"answer"} (the correct answer or interpretation for the question)
    \item \texttt{"selected\_passage"} (a snippet of the full poem, if relevant)
    \item \texttt{"full\_passage"} (the entire poem or the relevant portion of it, if needed)
    \item \texttt{"location"} (location reference in the poem, if relevant)
    \item \texttt{"r\_type"} (the rhetorical device type, if relevant)
    \item \texttt{"elements"} (the interpretation or content of the rhetorical device, if relevant)
    \item \texttt{"purpose"} (the purpose/effect of the rhetorical device, if relevant)
    \item \texttt{"ctype"} (the type of external context: historical, cultural, biographical, or literary, if relevant)
    \item \texttt{"celements"} (the factual external context details, if relevant)
    \item \texttt{"corresponding\_feature"} (the rhetorical device to which the context is connected, if relevant)
    \item \texttt{"pair"} (a short string describing the rhetorical device + external context pairing, if relevant)
\end{itemize}

\textbf{Your task:}
\begin{enumerate}
    \item \textbf{For each question object} in the input JSON, generate exactly \textbf{three (3) distractors}.
    \item Append those three distractors plus the correct answer (in the fourth position) to a new array \texttt{"choices"} within that same question object.
    \item \textbf{Do not} add any additional commentary or fields; only add \texttt{"choices"} to each question object, containing \texttt{[distractor1, distractor2, distractor3, correctAnswer]}.
\end{enumerate}

\textbf{The way you generate the three distractors depends on \texttt{question\_number}:}

\textbf{Q1}  

Here is a snippet of a poem: \texttt{\{selected\_passage\}}, selected for literary analysis from the full poem: \texttt{\{full\_passage\}}. One interpretation of how this selected passage stands out from the full work is: \texttt{\{answer\}}. I want three one-sentence distractors in the context of how this passage might differ from or connect to the rest of the poem. Number them 1, 2, 3, and do not provide other commentary.

\textbf{Q2}  

Here is a snippet of a poem: \texttt{\{selected\_passage\}}. In \texttt{\{location\}}, there is a \texttt{\{r\_type\}}. I want three rhetorical devices as distractors that might also appear in \texttt{\{location\}}. Number them 1, 2, 3, and do not provide other commentary.

\textbf{Q4}  

Here is a snippet of a poem: \texttt{\{selected\_passage\}}. In \texttt{\{location\}}, there's a \texttt{\{r\_type\}}. One interpretation is: \texttt{\{elements\}}. I want three one-sentence distractors for this interpretation. Number them 1, 2, 3, and do not provide other commentary.

\textbf{Q5}  

Here is a snippet of a poem: \texttt{\{selected\_passage\}}. In the \texttt{\{r\_type\}} that occurs in \texttt{\{location\}}, \texttt{\{elements\}}. One interpretation of this device's purpose/effect is: \texttt{\{purpose\}}. I want three one-sentence distractors for that interpretation. Number them 1, 2, 3, and do not provide other commentary.

\textbf{Q7}  

Here is a snippet of a poem: \texttt{\{selected\_passage\}}. We know it is relevant to this external context: “\texttt{\{ctype\}}, \texttt{\{celements\}}.” I want three distractors for other possible contexts that might be relevant. Structure each distractor similarly as 'context\_type, context\_content.' Number them 1, 2, 3, and do not provide other commentary.

\textbf{Q9}  

Here is a snippet of a poem: \texttt{\{selected\_passage\}}. It uses \texttt{\{corresponding\_feature\}}, and is relevant to this \texttt{\{ctype\}} context: \texttt{\{celements\}}. One interpretation for how the rhetorical device connects with that context is: \texttt{\{answer\}}. I want three one-sentence distractors for that interpretation. Number them 1, 2, 3, and do not provide other commentary.

\textbf{Q11}  

Here is the full poem: \texttt{\{full\_passage\}}. It contains a connection between a rhetorical device and an external context: \texttt{\{pair\}}. One interpretive argument for that connection is: \texttt{\{answer\}}. I want three distractors for that argument. Number them 1, 2, 3, and do not provide other commentary.

\textbf{Any other \texttt{question\_number}}  

Here is a snippet of a poem: \texttt{\{selected\_passage\}}. We have a question, and the correct answer is \texttt{\{answer\}}. I want three one-sentence distractors relevant to this question. Number them 1, 2, 3, and do not provide other commentary.

\textbf{Final Output:}  

After generating these three distractors for each question, insert them plus the correct answer into a new array field \texttt{"choices"} (with the correct answer as the last item) for each question object.

Finally, output the entire updated JSON array of questions, where each question has the form:
\begin{verbatim}
{
  "question_number": "...",
  "answer": "...",
  ...
  "choices": ["distractor1", "distractor2", "distractor3", "correct answer"]
}
\end{verbatim}

\textbf{Now proceed with distractor generation for the following JSON array of questions:}
\begin{quote}
[[QUESTIONS\_JSON]]
\end{quote}
\end{tcolorbox}

\subsection{Evaluation Prompts}
The following prompt is used for the evaluation of LLMs, reported in Table \ref{tab:main-results}:

\begin{tcolorbox}[colback=gray!10, colframe=black, boxrule=0.5pt]
\scriptsize
\textbf{Consider the following literary passage:}
\begin{quote}
[[PASSAGE]]
\end{quote}

\textbf{Question:}
\begin{quote}
[[QUESTION]]
\end{quote}

\textbf{Select the correct answer from the following choices:}
\begin{quote}
[[CHOICES]]
\end{quote}

\textbf{Response Format:}  

You must answer the question using JSON format, with the following schema:
\begin{verbatim}
{"answer": "A" | "B" | "C" | "D"}
\end{verbatim}

You should not include any other text in your response.
\end{tcolorbox}

\section{Details on Data Source}
\subsection{Exams}
\begin{figure*}
    \centering
    \includegraphics[width=1\linewidth]{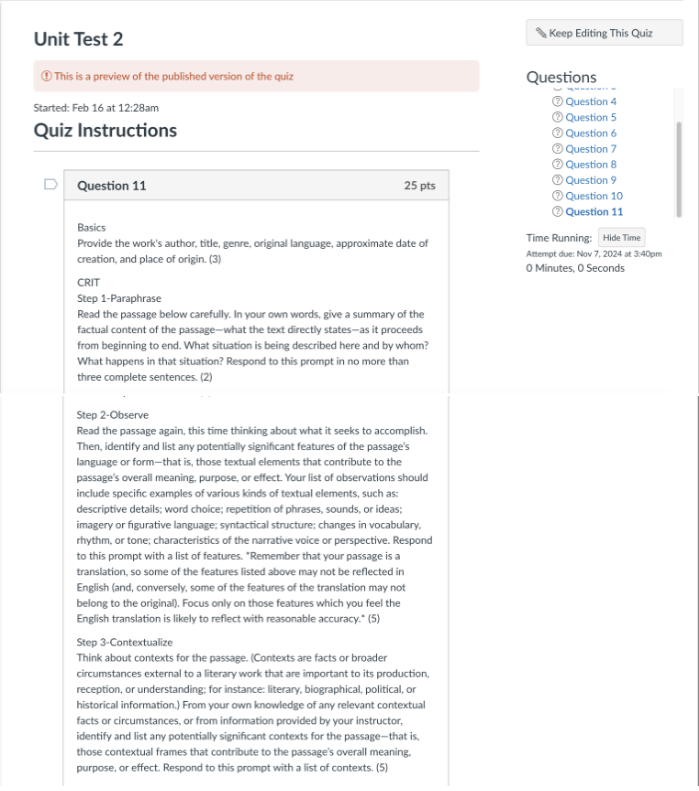}
    \caption{Exam interface (using unit test 2 as an example).}
    \label{fig:exam-interface}
\end{figure*}

Please see Figure \ref{fig:exam-interface}.
\subsection{Grading Rubrics}
\begin{figure*}
    \centering
    \includegraphics[width=1\linewidth]{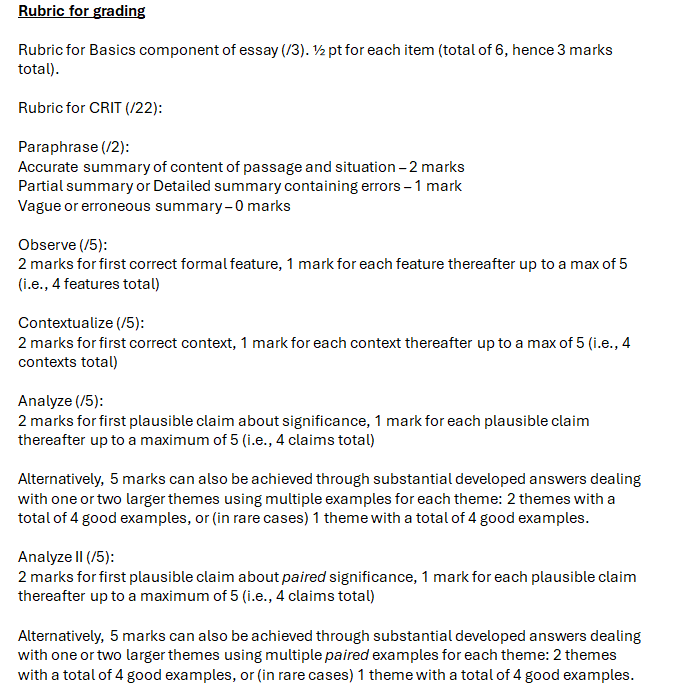}
    \caption{Rubric used to grade CRIT essays.}
    \label{fig:rubric}
\end{figure*}

Please see Figure \ref{fig:rubric}.

\section{Details on Human Evaluation}
\begin{figure*}
    \centering
    \includegraphics[width=1\linewidth]{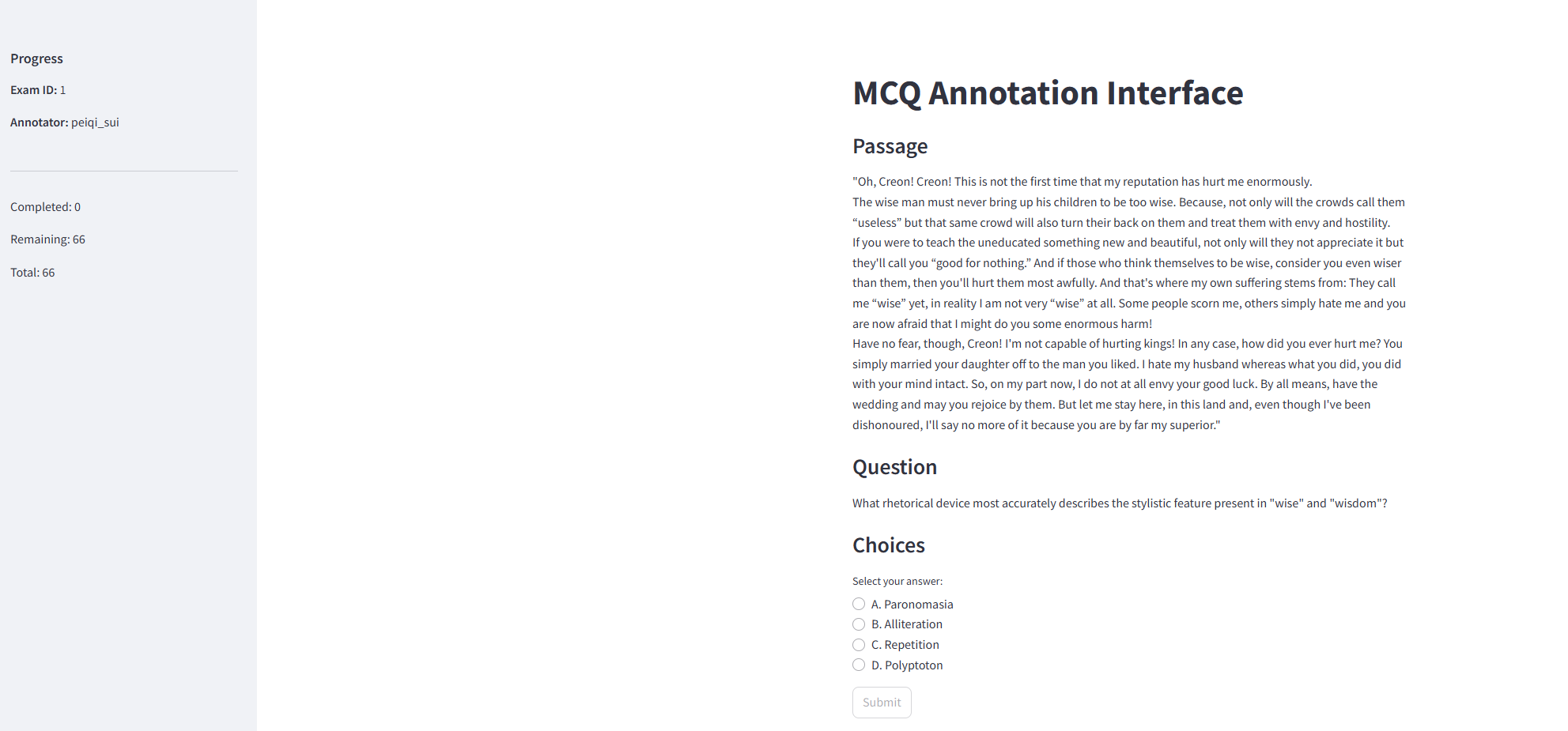}
    \caption{The online interface used for collecting the performance of evaluators on a subset of KRISTEVA to create the human baseline.}
    \label{fig:human-eval-interface}
\end{figure*}

Please see Figure \ref{fig:human-eval-interface} for the interface we used to perform the human evaluation.

\end{document}